\documentclass[10pt,twocolumn,letterpaper]{article}
\usepackage[pagenumbers]{wacv}

\usepackage{graphicx}
\usepackage{amsmath}
\usepackage{amssymb}
\usepackage{booktabs}

\usepackage{algorithm}
\usepackage{algorithmic}

\usepackage{amsmath}
\usepackage{amssymb}
\usepackage{bbm}

\usepackage{booktabs}
\usepackage{multirow}
\usepackage{makecell}

\usepackage[pagebackref,breaklinks,colorlinks]{hyperref}

\usepackage[capitalize]{cleveref}
\crefname{section}{Sec.}{Secs.}
\Crefname{section}{Section}{Sections}
\Crefname{table}{Table}{Tables}
\crefname{table}{Tab.}{Tabs.}

\newcommand\blfootnote[1]{
  \begingroup
  \renewcommand\thefootnote{}\footnote{#1}
  \addtocounter{footnote}{-1}
  \endgroup
}

\def\wacvPaperID{1255}
\def\confName{WACV}
\def\confYear{2024}

\newcommand{\BEST}[1]{\textbf{\textcolor[rgb]{1.00,0.00,0.00}{#1}}}
\newcommand{\SBEST}[1]{\textbf{\textcolor[rgb]{0.00,0.00,1.00}{#1}}}
\newcommand{\TBEST}[1]{\textbf{\textcolor[rgb]{1.00,0.00,1.00}{#1}}}

\begin{document}
\title{Zero-Shot Edge Detection with SCESAME: Spectral Clustering-based Ensemble for Segment Anything Model Estimation}

\author{
Hiroaki Yamagiwa$^{1,2}$
\qquad Yusuke Takase$^{1}$
\qquad Hiroyuki Kambe$^{2}$
\qquad Ryosuke Nakamoto$^{1,2}$\\
$^{1}$Kyoto University \qquad $^{2}$Rist Inc.\\
{\tt\small hiroaki.yamagiwa@sys.i.kyoto-u.ac.jp},
{\tt\small y.takase@sys.i.kyoto-u.ac.jp},\\
{\tt\small \{hiroyuki.kambe, ryosuke.nakamoto\}@rist.co.jp}
}
\maketitle

\begin{abstract}
This paper proposes a novel zero-shot edge detection with SCESAME, which stands for {\bf S}pectral {\bf C}lustering-based {\bf E}nsemble for {\bf S}egment {\bf A}nything {\bf M}odel {\bf E}stimation, based on the recently proposed Segment Anything Model (SAM). SAM is a foundation model for segmentation tasks, and one of the interesting applications of SAM is Automatic Mask Generation (AMG), which generates zero-shot segmentation masks of an entire image. AMG can be applied to edge detection, but suffers from the problem of overdetecting edges. Edge detection with SCESAME overcomes this problem by three steps: (1) eliminating small generated masks, (2) combining masks by spectral clustering, taking into account mask positions and overlaps, and (3) removing artifacts after edge detection. We performed edge detection experiments on two datasets, BSDS500 and NYUDv2. Although our zero-shot approach is simple, the experimental results on BSDS500 showed almost identical performance to human performance and CNN-based methods from seven years ago. In the NYUDv2 experiments, it performed almost as well as recent CNN-based methods. These results indicate that our method effectively enhances the utility of SAM and can be a new direction in zero-shot edge detection methods.
\blfootnote{Our code is available at \url{https://github.com/ymgw55/SCESAME}.}
\end{abstract}

\section{Introduction}
\begin{figure}[t]
    \centering
    \includegraphics[width=\columnwidth]{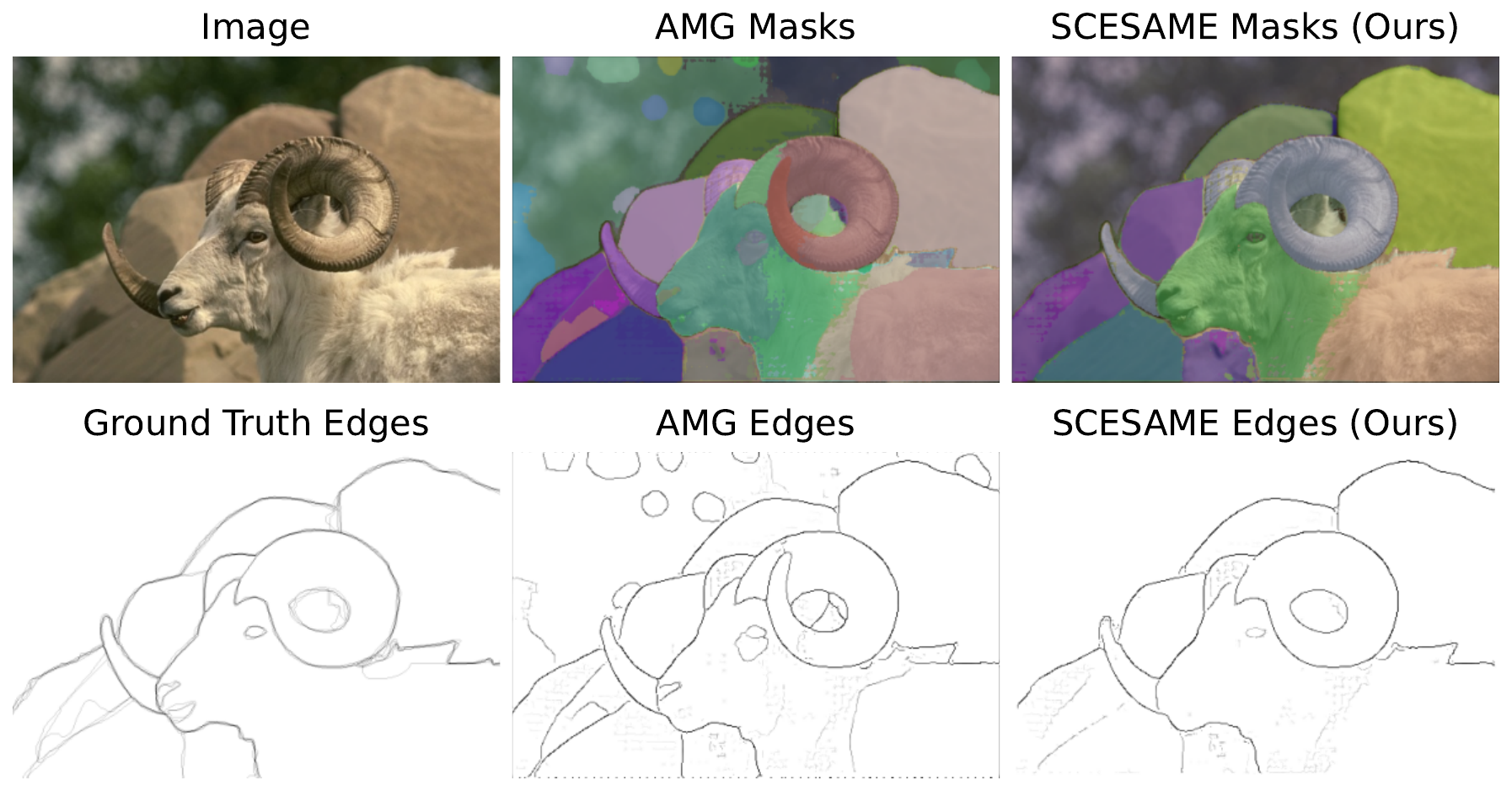}
    \caption{(upper row) Original image and masks generated by AMG and SCESAME. While AMG genarates 54 masks, SCESAME combines them into 9 masks after removing smaller ones. (lower row) Ground truth edges and edges from the masks generated by AMG and SCESAME. Unlike AMG, which excessively detects edges from background and shadows, SCESAME restricts such edge detection.}
    \label{fig:thumbnail}
\end{figure}
Foundation model~\cite{DBLP:journals/corr/abs-2108-07258} is the model that is pretrained on large-scale datasets and can be applied directly to downstream tasks, saving significant time and resources by eliminating the need to retrain the model for each specific task.

In the field of computer vision, several foundation models have been proposed for different tasks~\cite{DBLP:conf/iclr/DosovitskiyB0WZ21,DBLP:conf/icml/RadfordKHRGASAM21,DBLP:journals/corr/abs-2204-06125,DBLP:journals/corr/abs-2304-02643,zou2023segment,gpt4v}. 
The recently proposed Segment Anything Model (SAM)~\cite{DBLP:journals/corr/abs-2304-02643} is a foundation model for segmentation tasks, capable of generating segmentation masks from different types of few-shot prompts, including points, bounding boxes, and segmentations. An interesting application of SAM is Automatic Mask Generation (AMG), which generates zero-shot segmentation masks of an entire image. This approach involves providing SAM with a regular grid of points as prompts for an input image, predicting a segmentation mask for each point, and generating the segmentation for the entire image.

One application of AMG is edge detection, one of the most important tasks in image processing and computer vision, which involves identifying the boundaries or other significant features within an image~\cite{DBLP:conf/cvpr/PuHLGL22}. 
Edge detection is known to be applicable to downstream tasks such as segmentation~\cite{DBLP:journals/tip/LiuHC20,DBLP:conf/eccv/ChengWH020,DBLP:conf/nips/YuHBLLBAK21} and object detection~\cite{DBLP:conf/cvpr/CaellesMPLCG17,DBLP:conf/cvpr/QinZHGDJ19,DBLP:conf/aaai/ZhuL0YLCWQ22}.
Therefore, when zero-shot edge detection with AMG shows strong performance, it seems applicable to various downstream tasks. In practice, however, it tends to overdetect edges~\cite{DBLP:journals/corr/abs-2304-02643}, which is a significant problem.

Our motivation is to address such issue and propose a more effective zero-shot edge detection method based on AMG. To achieve this,  we focused on spectral clustering, a method that uses the spectral information (eigenvalues and eigenvectors) of a graph with affinities between data as edges, considering the data as points in a new space, and then clustering in that space. 

In this paper, we demonstrate a performance improvement in zero-shot edge detection with AMG by (1) removing smaller masks generated by AMG, (2) appropriately combing the remaining masks using spectral clustering, taking into account the mask positions and overlaps, and (3) eliminating artifacts that occur when edges are generated from masks. We refer to our method, which defines the affinity between masks generated by AMG and combines these masks using spectral clustering, as SCESAME: {\bf S}pectral {\bf C}lustering-based {\bf E}nsemble for {\bf S}egment {\bf A}nything {\bf M}odel {\bf E}stimation. 
Figure~\ref{fig:thumbnail} shows an example of masks and edge detection with AMG and SCESAME. While the AMG masks detect excess edges in the background and shadows, SCESAME remove small masks and effectively combine similar masks to reduce such edges.

Through experiments on BSDS500~\cite{DBLP:journals/pami/ArbelaezMFM11} and NYUDv2~\cite{DBLP:conf/eccv/SilbermanHKF12}, we found that our method exhibits performance nearly equivalent to human performance and CNN-based methods from seven years ago for BSDS500, and nearly equivalent to recent CNN-based methods for NYUDv2, despite being a simple zero-shot technique. 
While there is still a significant gap between edge detection with SCESAME and the state-of-the-art (SOTA) approaches, these results indicate that our method effectively enhances the utility of SAM and can be a new direction in zero-shot edge detection methods.
\section{Related Work}

\subsection{Edge Detection Method}
Edge detection has a long history, with many traditional methods proposed before the advent of deep learning-based methods. In particular, the Sobel filter~\cite{Kittler1983OnTA} is one of the earliest edge detection methods, with several advancements including the Canny method~\cite{DBLP:journals/pami/Canny86a}. 
In addition, Felz-Hut~\cite{DBLP:journals/ijcv/FelzenszwalbH04} achieves refined edge detection by comparing differences between regions using a graph-based representation.

In recent years, deep learning approaches to edge detection have been introduced, including methods using Convolutional Neural Networks (CNN)~\cite{DBLP:conf/iccv/XieT15,DBLP:journals/corr/Kokkinos15} and the Vision Transformer~\cite{DBLP:conf/cvpr/PuHLGL22}. Loss functions are also proposed to account for ambiguities in annotations~\cite{DBLP:journals/corr/Kokkinos15,Zhou_2023_CVPR}.

\subsection{SAM-based Model}
SAM generates segmentation masks with few prompts, so several segmentation models using SAM have been proposed. PerSAM~\cite{DBLP:journals/corr/abs-2305-03048} is a model that can segment specific concepts by one-shot tuning using a pair of an image and a mask. SAA+\cite{DBLP:journals/corr/abs-2305-10724} is a zero-shot anomaly detection model that uses Grounding DINO\cite{DBLP:journals/corr/abs-2303-05499} to generate bounding boxes from text and then provides them as prompts to SAM. 
Track Anything~\cite{DBLP:journals/corr/abs-2304-11968} is a model that can track objects in a video with just a few clicks. 
HQ-SAM~\cite{ke2023segment} is a model that performs additional learning for SAM parameters to generate more accurate masks. 

\subsection{Segmentation Method by Spectral Clustering}
Many methods have been proposed for segmentation using spectral clustering. 
The method that combines a blockwise segmentation strategy with stochastic ensemble consensus~\cite{DBLP:journals/pr/TungWC10} considers segment-level clustering and is related to our proposed approach. 
Linear spectral clustering~\cite{DBLP:conf/cvpr/LiC15} is a superpixel segmentation algorithm based on $k$-means clustering. 
The method of coupling local brightness, color, and texture cues using spectral clustering to detect contours has been proposed~\cite{DBLP:journals/pami/ArbelaezMFM11}.
For unsupervised semantic segmentation, a parametric approach has been proposed that employs neural network-based eigenfunctions to generate embeddings for spectral clustering~\cite{DBLP:journals/corr/abs-2304-02841}.
In the field of medical image segmentation, spectral clustering-based methods have been proposed by using prior information~\cite{DBLP:journals/mms/XiaGZ20} or by identifying the tumor region~\cite{DBLP:journals/jksucis/MaruthamuthuG20}.

\section{Zero-Shot Edge Detection with AMG}\label{sec:mask2edge}
\begin{figure}[t]
    \centering
    \includegraphics[width=0.95\columnwidth]{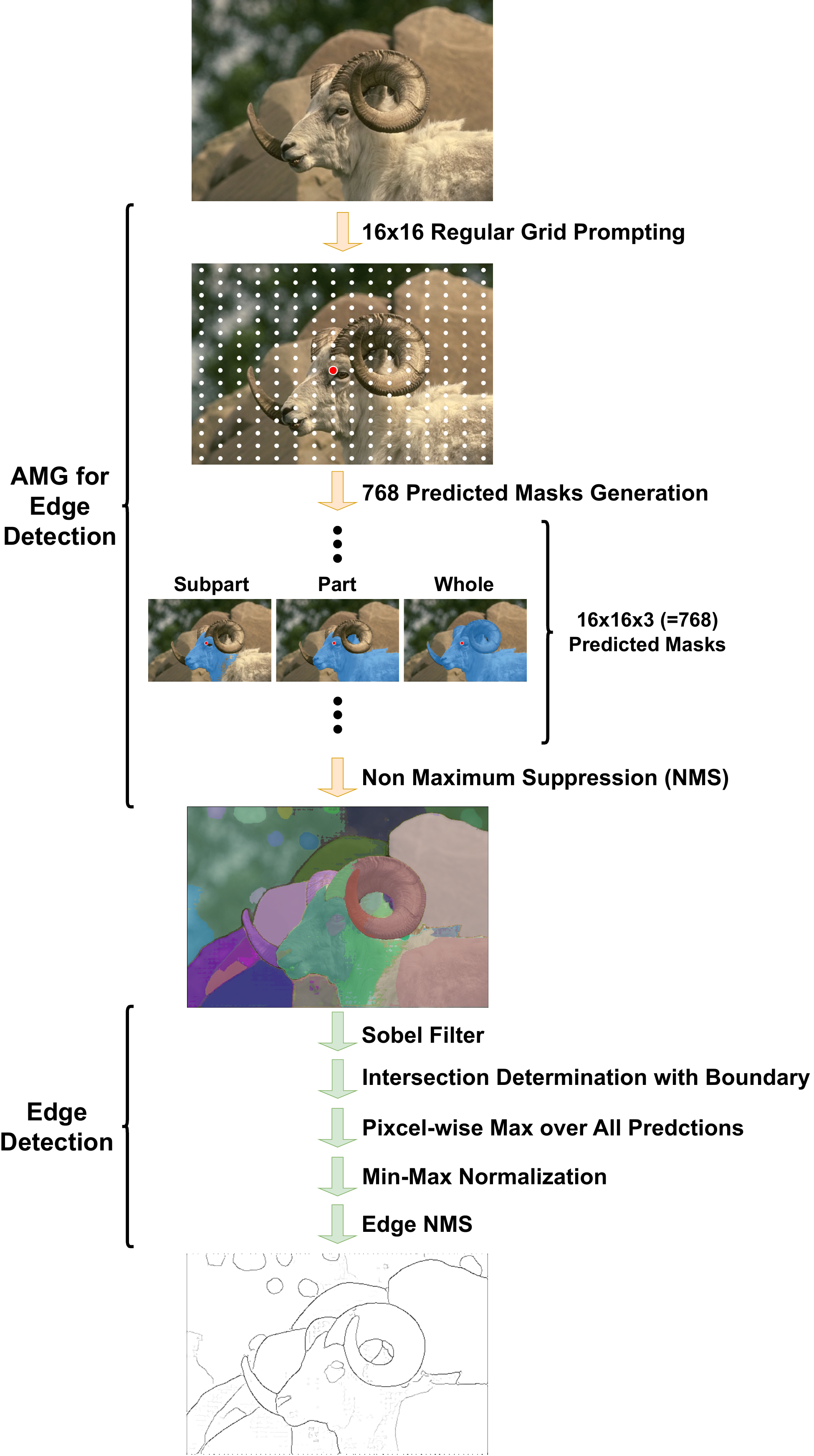}
    \caption{The zero-shot edge detection pipeline using AMG.}
    \label{fig:amg}
\end{figure}
In this section, we introduce the zero-shot edge detection pipeline using AMG, based on the original SAM paper~\cite{DBLP:journals/corr/abs-2305-03048}. Note that AMG for edge detection differs from standard AMG in terms of the number of points provided as prompts and the mask removal process, but we simply refer to AMG for edge detection as AMG throughout this paper unless there is confusion. For details on standard AMG, see the original SAM paper~\cite{DBLP:journals/corr/abs-2305-03048}.

First, we explain AMG for edge detection. A $16\times16$ regular grid of points is given to SAM as prompts, which predicts three different scale masks at each point, generating a total of 768 masks. Non-Maximum Suppression (NMS) is then applied to the masks to remove redundant masks.

Next, we explain the edge detection process using the AMG masks.
The logits of the masks are converted to probability values using an element-wise sigmoid function, and then a Sobel filter is applied for edge detection. 
During this process, all values except those at the boundaries of the masks are set to 0. 
Using the probabilities obtained for each mask, the maximum probability for each pixel over all the probabilities is determined, followed by min-max normalization over the entire image. 

Finally, a Gaussian blur is applied, and then edge NMS~\cite{DBLP:journals/pami/Canny86a,DBLP:journals/pami/DollarZ15} is used to thin the edges, although the Gaussian blur for improving edge NMS is not mentioned in the original paper.

Figure~\ref{fig:amg} illustrates the generation of masks by AMG and the edge detection process.
 It can be seen that in AMG, masks are generated at three scales: subpart, part, and whole, using a one point prompt. By performing edge detection on the masks remaining after NMS, edges are generated that reflect the contours of the masks. For details on the implementation, see \S~\ref{sec:impdetail}.

\section{Spectral Clustering}~\label{sec:SC}
\begin{algorithm}[t]
\caption{Spectral Clustering}
\begin{algorithmic}[1]
\REQUIRE Affinity matrix $\mathbf{W}=(w_{ij}) \in \mathbb{R}_{\geq 0}^{n\times n}$, number of clusters $k$.
\ENSURE Clusters $A_1,...,A_k$, $A_i = \{j \,|\, \mathbf{y}_j \in C_i\}$.
\STATE Compute the graph Laplacian $\mathbf{L}$.
\STATE Select the $k$ smallest eigenvalues of $\mathbf{L}$ and denote their corresponding eigenvectors as $\mathbf{u}_1,...,\mathbf{u}_k$.
\STATE Define the matrix $\mathbf{U}=[\mathbf{u}_1, \cdots, \mathbf{u}_k]\in\mathbb{R}^{n\times k}$, and let the row vectors of $\mathbf{U}$ be $\mathbf{y}_i \in \mathbb{R}^k$. Then, $\mathbf{U}=[\mathbf{y}_1,\cdots,\mathbf{y}_n]^\top$.
\STATE Treat the vectors $(\mathbf{y}_i)_{i=1}^n$ as corresponding to each vertex, and use the $k$-means clustering to classify them into clusters $C_1,...,C_k$.
\end{algorithmic}
\label{algo:SC}
\end{algorithm}
\begin{figure}[t!]
    \centering
    \includegraphics[width=0.85\columnwidth]{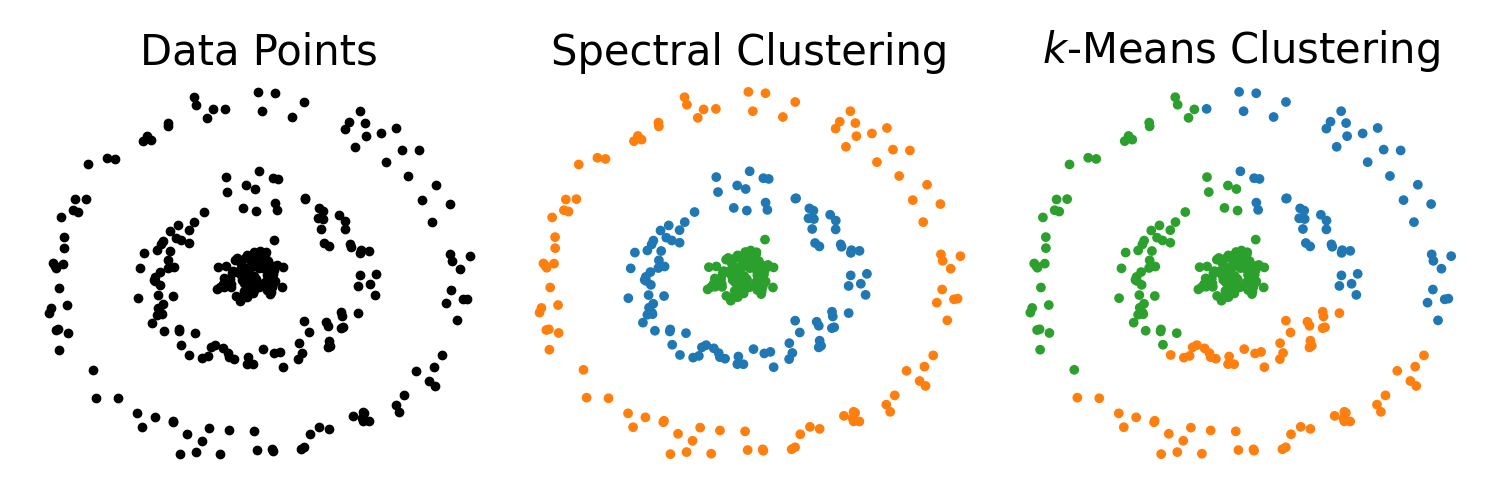}
    \caption{The utility of spectral clustering: (left) Inspired by a simple example~\cite{DBLP:conf/nips/Zelnik-ManorP04}, 100 points are randomly sampled from circles of radius 0.1, 0.5, and 1.0, and Gaussian noise is added. (middle) An affinity matrix is computed from a 10-nearest neighbor graph, and normalized spectral clustering with $k=3$ classifies the three circles separately. (right) With $k$-means clustering, where $k=3$, the points are classified based on their distance proximity only, and the three circles cannot be classified separately.}
    \label{fig:sc_sample}
\end{figure}
In this section, based on the well-known tutorial on spectral clustering~\cite{DBLP:journals/sac/Luxburg07}, we explain the algorithm for spectral clustering~\cite{DBLP:conf/nips/NgJW01} that is used in our proposed method.

Consider an undirected graph $G$ with vertex set $V=\{v_1,\cdots,v_n\}$, and define an affinity matrix between vertices $\mathbf{W}=(w_{ij})\in \mathbb{R}_{\geq 0}^{n\times n}$. Also, define the degree matrix $\mathbf{D}=\mathrm{diag}(d_1,\cdots,d_n)\in \mathbb{R}^{n\times n}$ where $d_i:=\sum_{j=1}^n w_{ij}$. 

The graph Laplacian for $\mathbf{D}, \mathbf{W}$ is defined as:
\begin{align}
    \mathbf{L} := \mathbf{D} - \mathbf{W}\in\mathbb{R}^{n\times n}.
\end{align}
Since $\mathbf{D}, \mathbf{W}$ are symmetric matrices, $\mathbf{L}$ is also symmetric. For any $\mathbf{f}=(f_1,\cdots,f_n)^\top\in\mathbb{R}^{n}$,
\begin{align}
    \mathbf{f}^\top \mathbf{L} \mathbf{f} = \frac{1}{2}\sum_{i,j=1}^n w_{ij}(f_i - f_j)^2 \geq 0
\end{align}
holds, so $\mathbf{L}$ is a positive semidefinite matrix~\cite{DBLP:journals/sac/Luxburg07}. 
By defining a constant vector with all components equal to $1$ as $\mathbbm{1}\in\mathbb{R}^n$, it follows from the definitions of $\mathbf{D}, \mathbf{W}$ that $\mathbf{L}\mathbbm{1}=(\mathbf{D}-\mathbf{W})\mathbbm{1}=\mathbf{0}$. Thus, the smallest eigenvalue of $\mathbf{L}$ is $0$, and its corresponding eigenvector is $\mathbbm{1}$.

The multiplicity $k$ of the eigenvalue $0$ of $\mathbf{L}$ corresponds to the number of connected components in $G$, and if we denote their index sets as $A_1, \cdots, A_k$, the eigenvectors are given by the indicator vectors $\mathbbm{1}_{A_1}, \cdots, \mathbbm{1}_{A_k}\in\mathbb{R}^n$~\cite{DBLP:journals/sac/Luxburg07}. Here, for $\mathbbm{1}_{A_l}=(g_1,\cdots,g_n)^\top$, $g_i=1$ if $i\in A_l$, and $g_i=0$ otherwise.

Next, we define a matrix $\mathbf{U}=[\mathbbm{1}_{A_1}, \cdots, \mathbbm{1}_{A_k}]\in\mathbb{R}^{n\times k}$ with column vectors $\mathbbm{1}_{A_1}, \cdots,  \mathbbm{1}_{A_k}$. We denote the row vectors of $\mathbf{U}$ as $\mathbf{y}_i\in\mathbb{R}^{k}$, that is $\mathbf{U}=[\mathbf{y}_1,\cdots,\mathbf{y}_n]^\top$. 
Here, vertex $v_i$ belongs to the connected component corresponding to the index where the component of $\mathbf{y}_i$ is equal to $1$.
However, the graph does not necessarily have $k$ connected components. 
In such cases, by considering $k$ smallest eigenvalues of $\mathbf{L}$ and the eigenvectors instead of $\mathbbm{1}_{A_1}, \cdots, \mathbbm{1}_{A_k}$, we can redefine a matrix $\mathbf{U}$ with these eigenvectors and use $k$-means clustering to determine the cluster to which the row vector $\mathbf{y}_i$ corresponding to the vertex $v_i$ belongs. 
The algorithm is described in Algorithm~\ref{algo:SC}.
For details, see the tutorial~\cite{DBLP:journals/sac/Luxburg07}.

The matrix $\mathbf{L}$ is precisely referred to as the unnormalized graph Laplacian, and the normalized graph Laplacian $\mathbf{L}_{\mathrm{sym}}$ is defined using the identity matrix $\mathbf{I}\in\mathbb{R}^{n\times n}$ as follows:
\begin{align}
    \mathbf{L}_{\mathrm{sym}}:=\mathbf{I}-\mathbf{D}^{-1/2}\mathbf{W}\mathbf{D}^{-1/2}\in\mathbb{R}^{n\times n}
\end{align}
Normalized spectral clustering can be considered using the same procedure as in the unnormalized case.

A comparison between normalized spectral clustering and $k$-means clustering for two-dimensional points is shown in Figure~\ref{fig:sc_sample}. Normalized spectral clustering can classify the three circles separately, while $k$-means clustering can not do so. Since spectral clustering also uses $k$-means clustering, this example illustrates that the row vectors $\mathbf{y}_i$ of the matrix $\mathbf{U}$ provide suitable embeddings.

\section{Proposed Method}\label{sec:proposed}
\begin{figure}[t]
    \centering
    \includegraphics[width=\columnwidth]{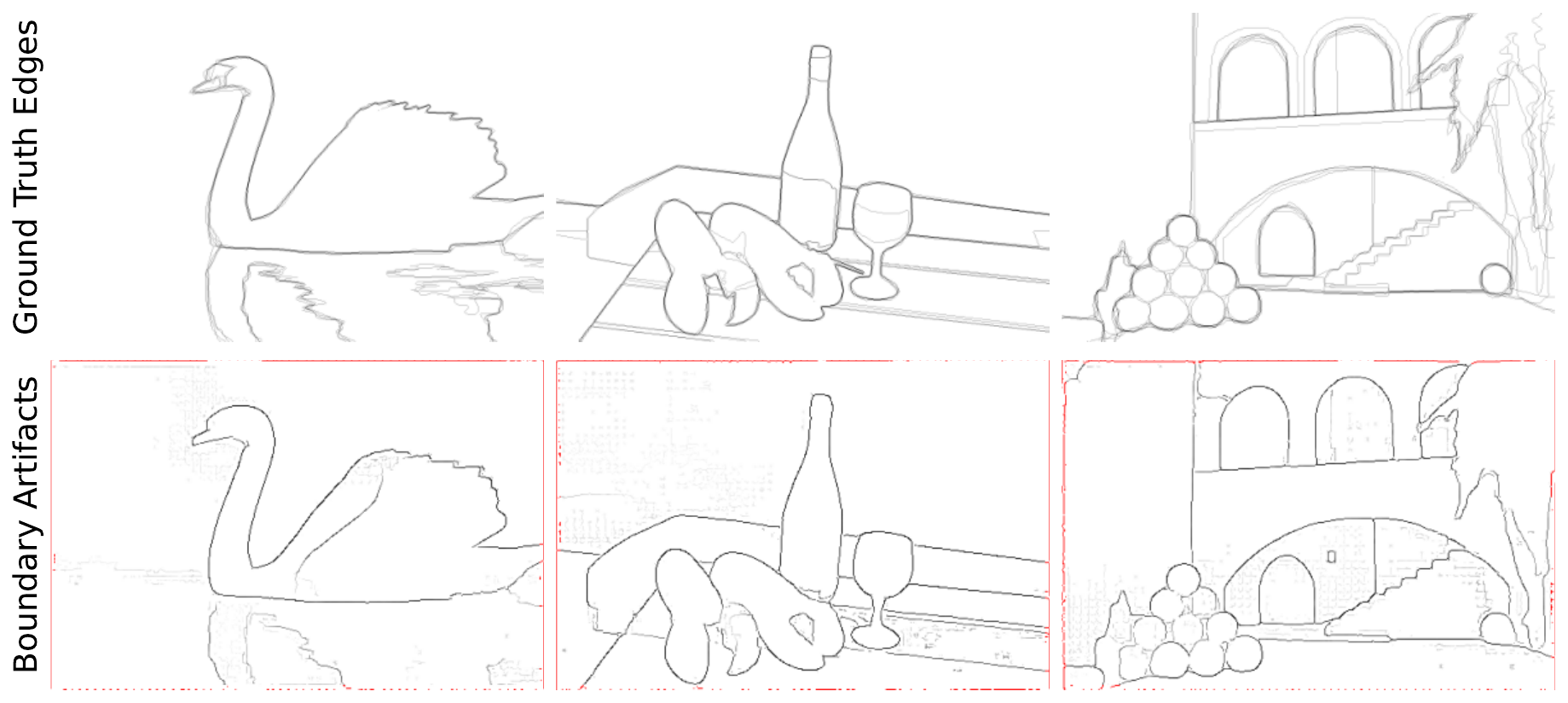}
    \caption{(upper row) Ground truth edges. (lower row) The edges from the SCESAME masks, with edges within 5 pixels of the image boundary highlighted in red.}
    \label{fig:bzp}
\end{figure}
\begin{figure*}[t]
    \centering
    \includegraphics[width=0.925\textwidth]{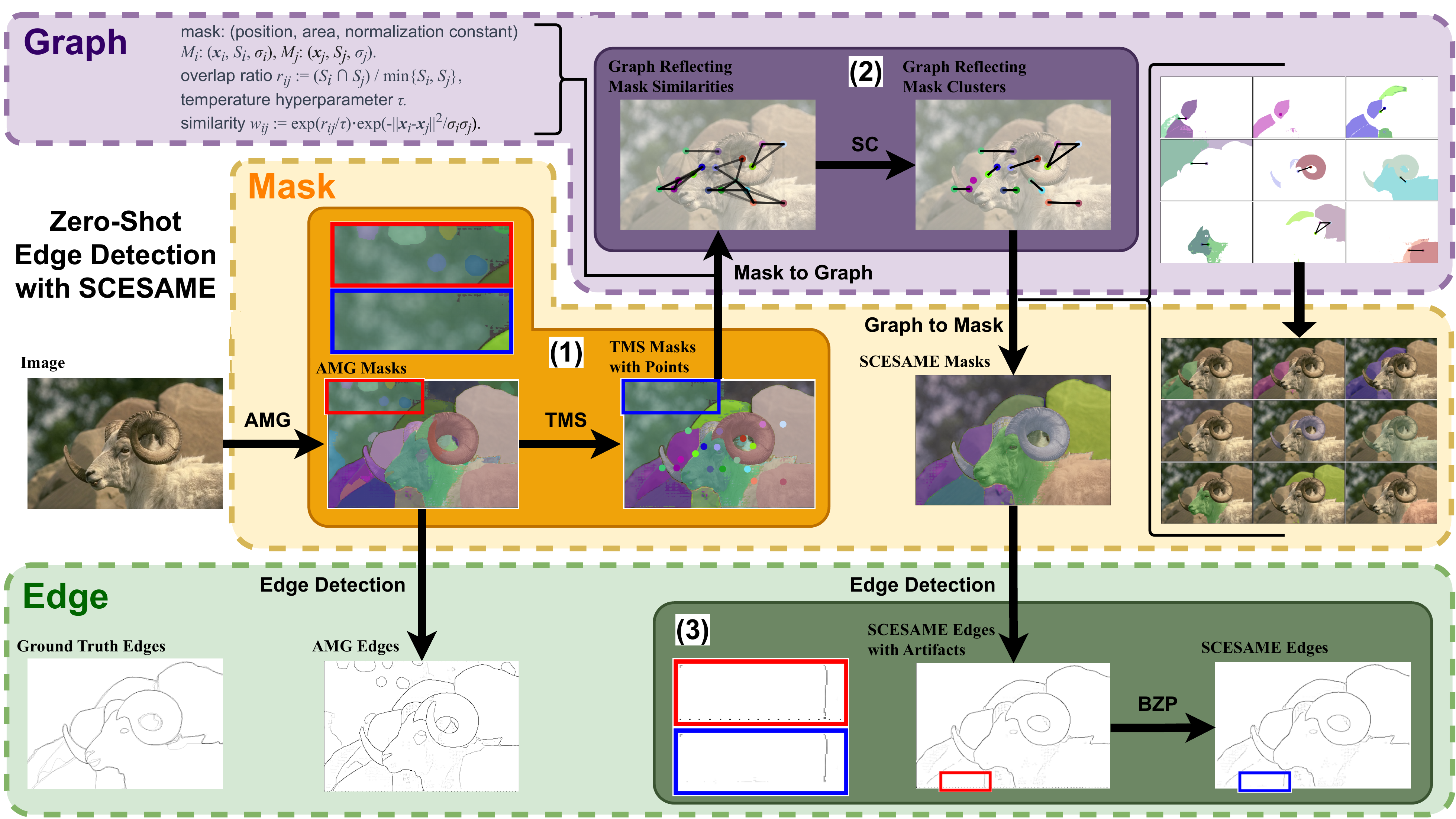}
    \caption{Representation of zero-shot edge detection with SCESAME for $t = 3, c = 2, p = 5$. For the AMG masks, (1) small masks are removed using TMS; (2) center points are assigned to each mask to define a graph, and SC is used to divide points into clusters, with masks corresponding to points in the same cluster combined to generate the SCESAME mask; (3) since the edges from the SCESAME masks contain boundary artifacts, BZP is used to remove them. In addition, in TMS and BZP, the changes before and after the process are magnified and shown in the red and blue regions, and in SC, all the points to be combined and their corresponding masks are shown.}
    \label{fig:method}
\end{figure*}
\begin{figure*}[t!]
    \centering
    \includegraphics[width=0.925\textwidth]{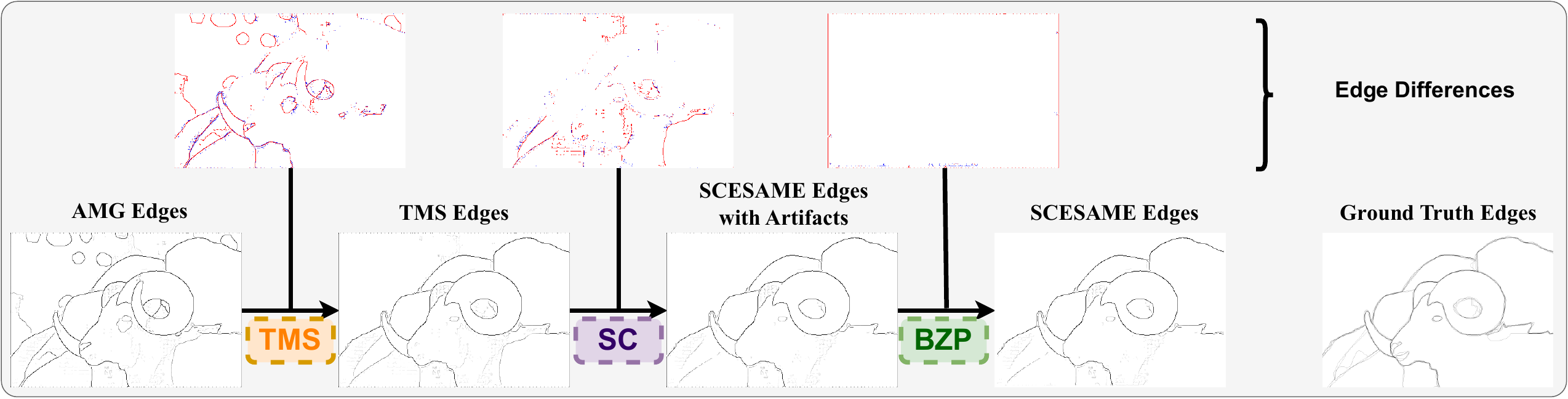}
    \caption{Differences in edges before and after processing in TMS, SC, BZP. Edges before processing are shown in red, and those after processing are shown in blue, with only differences greater than 0.05 shown for visibility.}
    \label{fig:method_diff}
\end{figure*}
\begin{figure*}[t!]
    \centering
    \includegraphics[width=0.99\textwidth]{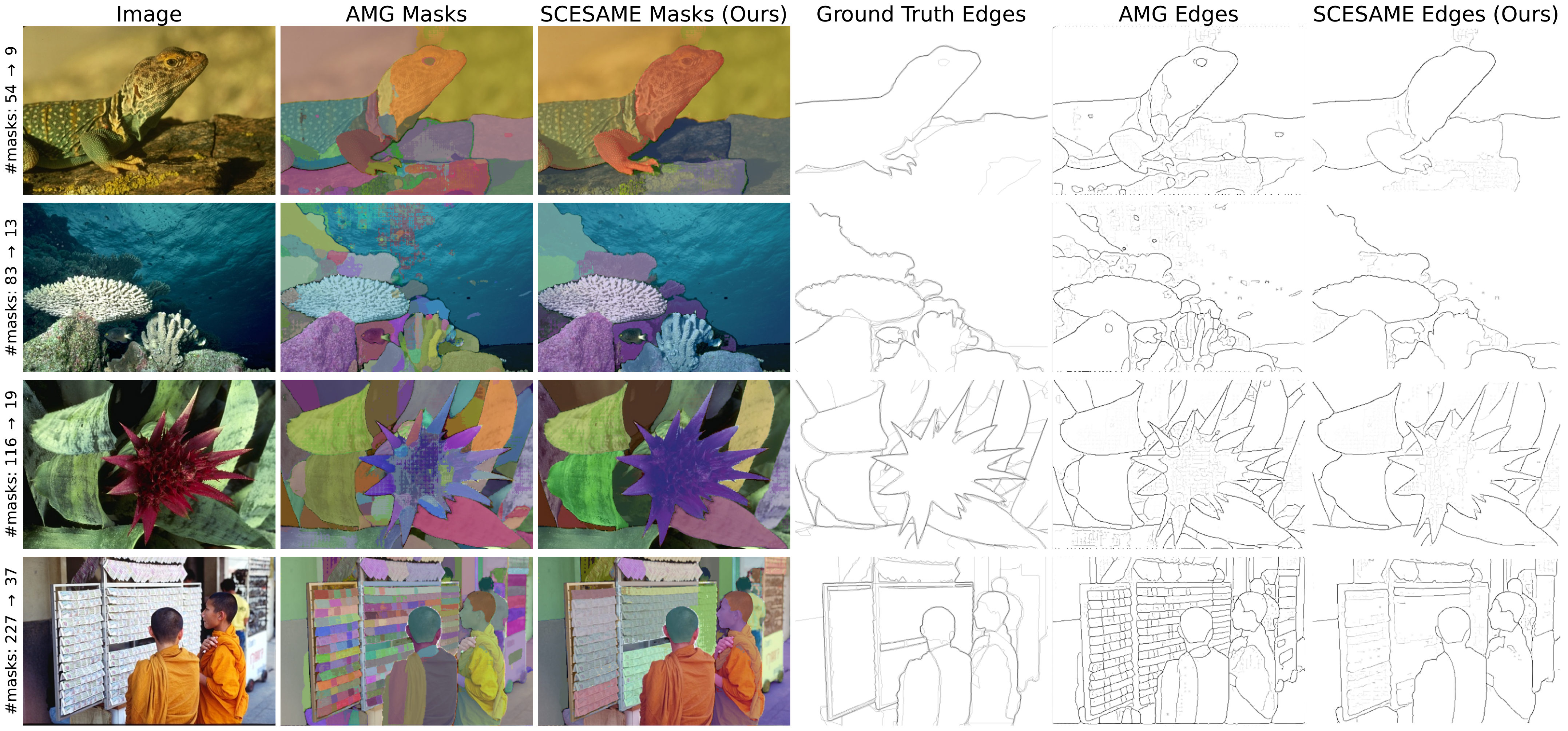}
    \caption{For some BSDS500~\cite{DBLP:journals/pami/ArbelaezMFM11} images, this figure shows the images, the masks generated by AMG and SCESAME, the ground truth edges, and the edges generated by AMG and SCESAME. It also shows the change in the number of masks between AMG and SCESAME.}
    \label{fig:sample}
\end{figure*}
This section describes three steps for edge detection with SCESAME: {\bf S}pectral {\bf C}lustering-based {\bf E}nsemble for {\bf S}egment {\bf A}nything {\bf M}odel {\bf E}stimation, based on AMG.

\subsection{Removal of Small Noise Masks}\label{sec:TMS}
Edge detection with AMG tends to be overly sensitive to minor changes that humans would not notice. 
For example, comparing the original image with the AMG masks in Figure~\ref{fig:thumbnail}, we can see that detail edges are detected in response to background light because AMG generates small masks. However, since humans do not detect edges for such small changes, edge detection with AMG results in excessive detection of unnecessary edges.

Based on this observation, we propose a preprocessing step to remove small masks that would act as noise during edge detection. We sort the AMG masks by size and select only the top $1/t$ masks, where $t\in \mathbb{N}$. We call this operation Top Mask Selection (TMS). Despite its simplicity, edge detection with TMS improves performance over edge detection with AMG. See \S~\ref{sec:ablation} for details.

In the next section, we show a method to achieve higher performance edge detection based on the TMS masks.

\subsection{Mask Ensemble Using Spectral Clustering}\label{sec:SC2}
TMS selects masks based solely on their size, without taking into account their positions or overlaps. This may lead to the overdetection of unnecessary edges. To manage this issue, we propose merging the masks obtained by TMS. We define an affinity based on mask positions and overlaps, and use Spectral Clustering (SC) for this merging process.

Let the masks obtained by TMS be $\{M_i\}_{i=1}^n$. For each mask $M_i$, the position of the mask $\mathbf{x}_i$ is determined by the center point of its bounding box. Let $S_i$ denote the area of $M_i$. If we define the overlapping area between $M_i$ and $M_j$ as $S_i \cap S_j$, then the ratio of this overlap  to the smaller mask area is $r_{ij} := S_i \cap S_j / \min\{S_i, S_j\} \in [0,1]$.

Using the ratio of the overlapping area $r_{ij}$ and the distance between the masks $\|\mathbf{x}_i-\mathbf{x}_j\|$, we model the affinity $w_{ij}$ between $M_i, M_j$ as follows:
\begin{align}
    w_{ij} := \exp{\left(\frac{r_{ij}}{\tau}\right)}\exp{\left(-\frac{\|\mathbf{x}_i-\mathbf{x}_j\|^2}{\sigma_i \sigma_j}\right)}\label{eq:scesame}
\end{align}
In the above, $\tau\in \mathbb{R}_{>0}$ is the temperature hyperparameter, and $\sigma_i\in \mathbb{R}_{>0}$ is a normalization constant specific to $\mathbf{x}_i$, determined by the distance to the seventh closest point\cite{DBLP:conf/nips/Zelnik-ManorP04}. According to the definition of (\ref{eq:scesame}), $w_{ij}$ increases as $r_{ij}$ increases and as $\|\mathbf{x}_i-\mathbf{x}_j\|$ decreases. When $\tau<1$, the affinity emphasize the ratio of the overlapping area rather than the distance between the masks.
Note that, if $r_{ij}=0$ in equation (\ref{eq:scesame}), it aligns with the local scaling affinity presented in~\cite{DBLP:conf/nips/Zelnik-ManorP04}.

From (\ref{eq:scesame}), a similarity matrix $\mathbf{W}=(w_{ij})\in\mathbb{R}_{\geq 0}^{n\times n}$ can be derived. First, we set the number of clusters to $k=\max\{\lfloor n/c\rfloor,2\}\in \mathbb{N}$, with $c\,(>1)\in \mathbb{N}$, where $\lfloor\cdot\rfloor$ is floor function. Then, we perform normalized spectral clustering on $\{\mathbf{x}_i\}_{i=1}^n$. Finally, we generate new masks $\{\tilde{M}_i\}_{i=1}^k$ by combining the masks associated with the same cluster.

We call the entire procedure of combining masks using spectral clustering, including TMS, SCESAME: {\bf S}pectral {\bf C}lustering-based {\bf E}nsemble for {\bf S}egment {\bf A}nything {\bf M}odel {\bf E}stimation. SCESAME is designed to generate zero-shot segmentation masks such as AMG. While edge detection can be done using the SCESAME masks $\{\tilde{M}_i\}_{i=1}^k$, there may be artifacts when extracting edges from these masks.  
The next section describes a method for dealing with these artifacts.

\subsection{Removal of Boundary Artifacts}\label{sec:BZP}
Methods like AMG and SCESAME segment an entire image, resulting in the appearance of mask boundaries at the image boundaries. Consequently, when detecting edges from the mask, the mask boundaries tend to be detected as artifacts at the image border. We refer to these unintended artifacts as boundary artifacts. Figure~\ref{fig:bzp} highlights in red the SCESAME edges within 5 pixels of the image boundary and compares them with the ground truth edges. We can see that boundary artifacts appear in the SCESAME edges, although they are not in the ground truth edges.

For this reason, when detecting edges from AMG or SCESAME masks, we introduce a post-processing step termed Boundary Zero Padding (BZP). In this process, we fill all pixels within $p$ pixels of the image boundary with zeros, where $p\in \mathbb{N}$. BZP step is applied after calculating the maximum probability for each pixel in the edge detection process. While there may be concerns about zero-padding potentially obscuring true positive edges and thereby degrading performance, our experiments demonstrate the high effectiveness of BZP. Further details can be found in \S~\ref{sec:ablation}.

\subsection{Zero-Shot Edge Detection with SCESAME}
In this paper, unless otherwise noted, zero-shot edge detection with SCESAME includes BZP processing. 
Figure~\ref{fig:method} illustrates how zero-shot edge detection with SCESAME is constructed from the procedures described in \S~\ref{sec:TMS}, \S~\ref{sec:SC2}, and \S~\ref{sec:BZP}.
We can see that small masks are removed by TMS, remaining masks are adaptively combined  by SC, and boundary artifacts are removed by BZP.

Figure~\ref{fig:method_diff} shows the edges generated by TMS, SC, and BZP, along with the differences between them. First, most of the edges from small masks are removed by TMS, followed by the removal of detail edges such as shadows by SC. Finally, boundary artifacts are removed by BZP.

Figure~\ref{fig:sample} shows examples of edge detection with AMG and SCESAME for some BSDS500~\cite{DBLP:journals/pami/ArbelaezMFM11} images.

\section{Experiments}
\begin{table}[t!]
\setlength{\abovecaptionskip}{0pt}
\centering
\footnotesize
\renewcommand\arraystretch{0.9}
\renewcommand\tabcolsep{5.0pt}
\begin{tabular}{c|l|c|ccc}
    \midrule
    \multicolumn{2}{c|}{Method}          &\makecell[c]{Pub.'Year}     & ODS       & OIS       & AP    \\
    \midrule
    & Human~\cite{DBLP:journals/corr/Kokkinos15}          & ICLR'16      & \SBEST{0.803} & -  & - \\
    \midrule
    \multirow{9}{*}{\rotatebox{90}{Traditional}}
    & Canny~\cite{DBLP:journals/pami/Canny86a}     & PAMI'86     & 0.600     & 0.640     & 0.580 \\
    & Felz-Hutt~\cite{DBLP:journals/ijcv/FelzenszwalbH04}   & IJCV'04      & 0.610     & 0.640     & 0.560 \\
    & gPb-owt-ucm~\cite{DBLP:journals/pami/ArbelaezMFM11}  & PAMI'10     & 0.726     & 0.757     & 0.696 \\
    & SCG~\cite{DBLP:conf/nips/RenB12}              & NeurIPS'12   & 0.739     & 0.758     & 0.773 \\
    & Sketch Tokens~\cite{DBLP:conf/cvpr/LimZD13}      & CVPR'13      & 0.727     & 0.746     & 0.780 \\
    & PMI~\cite{DBLP:conf/eccv/IsolaZKA14}                 & ECCV'14      & 0.741     & 0.769     & 0.799 \\
    & SE~\cite{DBLP:journals/pami/DollarZ15}                & PAMI'14     & 0.746     & 0.767     & 0.803 \\
    & OEF~\cite{DBLP:journals/corr/HallmanF14}               & CVPR'15      & 0.746     & 0.770     & 0.820 \\
    & MES~\cite{DBLP:conf/iccv/SironiLF15}                & ICCV'15      & 0.756     & 0.776     & 0.756 \\
    \midrule
    \multirow{10}{*}{\rotatebox{90}{7 to 8-Year-Old CNN}}
    & DeepEdge~\cite{DBLP:conf/cvpr/BertasiusST15}   & CVPR'15      & 0.753     & 0.772     & 0.807 \\
    & CSCNN~\cite{hwang2015pixelwise}             & ArXiv'15       & 0.756     & 0.775     & 0.798 \\
    & MSC~\cite{DBLP:journals/pami/SironiTLF16}                & PAMI'15     & 0.756     & 0.776     & 0.787 \\
    & DeepContour~\cite{DBLP:conf/cvpr/ShenWWBZ15}  & CVPR'15      & 0.757     & 0.776     & 0.800 \\
    & HFL~\cite{DBLP:conf/iccv/BertasiusST15}             & ICCV'15      & 0.767     & 0.788     & 0.795 \\
    & HED~\cite{DBLP:conf/iccv/XieT15}                   & ICCV'15      & 0.788     & 0.808     & \TBEST{0.840} \\
    & Deep Boundary~\cite{DBLP:journals/corr/Kokkinos15}& ICLR'16      & \BEST{0.813}     & \BEST{0.831}     & \BEST{0.866} \\
    & CEDN~\cite{DBLP:conf/cvpr/YangPCL016}                & CVPR'16      & 0.788     & 0.804     & - \\
    & RDS~\cite{DBLP:conf/cvpr/LiuL16}                   & CVPR'16      & 0.792     & 0.810     & 0.818 \\
    & COB~\cite{DBLP:conf/eccv/ManinisPAG16}               & ECCV'16      & 0.793     & \SBEST{0.820}     & \SBEST{0.859} \\
    \midrule
    \multirow{3}{*}{\rotatebox{90}{SAM}}
    & SAM~\cite{DBLP:journals/corr/abs-2304-02643}           & ICCV'23      & 0.768 & 0.786  & 0.794 \\
    & SAM~\cite{DBLP:journals/corr/abs-2304-02643} (Recalc.)          & ICCV'23      & 0.730 & 0.754  & 0.729 \\
    & SAM-p5 (Our Baseline)      & -   & 0.754 & 0.779  & 0.763 \\
    \midrule
    \multirow{4}{*}{\rotatebox{90}{Ours}}
    & SCESAME-t2c2p5           & \multirow{4}{*}{-} & 0.796 & 0.812 & 0.780 \\
    & SCESAME-t2c3p5          &       & 0.797 & 0.811 & 0.768 \\
    & SCESAME-t3c2p5           &       & \TBEST{0.800} & \TBEST{0.814} & 0.773 \\
    & SCESAME-t3c3p5           &       & 0.796 & 0.809 & 0.753 \\
    \midrule
    \midrule
    \multirow{4}{*}{\rotatebox{90}{SOTA}}
    & EDTER-MS~\cite{DBLP:conf/cvpr/PuHLGL22}           & CVPR'22      & 0.840 & 0.858  & 0.896 \\
    & EDTER-MS-VOC~\cite{DBLP:conf/cvpr/PuHLGL22}           & CVPR'22      & {\bf 0.848} & {\bf 0.865}  & 0.903 \\
    & UAED-MS~\cite{Zhou_2023_CVPR}           & CVPR'23      & 0.837 & 0.855 & 0.897 \\
    & UAED-MS-VOC~\cite{Zhou_2023_CVPR}           & CVPR'23     & 0.844  & 0.864 & {\bf 0.905} \\
    \midrule
\end{tabular}
\caption{Results on BSDS500~\cite{DBLP:journals/pami/ArbelaezMFM11} testing set. The notation t3c2p5 represents $t=3,c=2,p=5$ and so on. The best three results, excluding SOTA methods, are highlighted in \BEST{red}, \SBEST{blue}, and \TBEST{purple}. SOTA methods are highlighted in {\bf bold}. MS indicates multi-scale testing~\cite{DBLP:conf/cvpr/PuHLGL22,Zhou_2023_CVPR}, and VOC indicates training with additional PASCAL VOC data~\cite{DBLP:journals/ijcv/EveringhamGWWZ10}.}
\label{tab:bsds}
\vspace{-9pt}
\end{table}
\subsection{Datasets}
{\bf BSDS500}~\cite{DBLP:journals/pami/ArbelaezMFM11} consists of 500 RGB natural images, divided into 100 for training, 200 for validation, and 200 for testing. Each image was manually annotated by 4-9 annotators, with an average of 5 annotations per image.

{\bf NYUDv2}~\cite{DBLP:conf/eccv/SilbermanHKF12} contains 1449 indoor scenes consisting of RGB and HHA image pairs, divided into 381 for training, 414 for validation, and 654 for testing.

Since edge detection with SCESAME is a zero-shot technique designed for RGB images, we use only the test RGB images from both datasets.

\subsection{Implementation Details}\label{sec:impdetail}
Since the original implementation of edge detection with AMG used in the SAM paper~\cite{DBLP:journals/corr/abs-2304-02643} is not publicly available, we reimplemented it based on the description in the paper. Specifically, we set the NMS threshold to 0.7, determined the mask boundary using the Sobel filter, applied a Gaussian blur with kernel size 3 before edge NMS, and used OpenCV~\cite{opencv_library}'s Structured Forests~\cite{DBLP:journals/pami/DollarZ15} model\footnote{\url{https://github.com/opencv/opencv_extra/blob/master/testdata/cv/ximgproc/model.yml.gz}} for edge NMS\footnote{In the original paper, Canny edge NMS~\cite{DBLP:journals/pami/Canny86a} was used for edge NMS. However, in our environment, it did not produce the edges reported in the paper. This part needs further investigation and improvement.}.

For the value of $\tau$ in (\ref{eq:scesame}), as seen in \S~\ref{sec:SC2}, we set $\tau=0.5<1$ to emphasize the ratio of overlapping area between masks rather than their distance. For BZP, we set $p=5$, and the values of $t$ and $c$ used in SCESAME are discussed in \S~\ref{sec:results}. We use scikit-learn~\cite{scikit-learn} to perform normalized spectral clustering and Python implementation for prediction evaluation~\footnote{\url{https://github.com/Britefury/py-bsds500/}}.

\subsection{Evaluation Metric}
We use Optimal Dataset Scale (ODS), Optimal Image Scale (OIS), and Average Precision (AP)~\cite{DBLP:journals/corr/abs-2304-02643,DBLP:conf/cvpr/PuHLGL22} as evaluation metrics. ODS is the F-score when selecting the optimal threshold for the entire dataset, and OIS is the F-score when selecting the optimal threshold for each image, with thresholds ranging from 0.01 to 0.99. AP is the integrated value of the precision-recall curve. Following previous
works~\cite{DBLP:conf/iccv/XieT15,DBLP:journals/corr/LiuCHWB16,DBLP:conf/cvpr/PuHLGL22}, the localization tolerance is set to 0.0075 for BSDS500 and 0.011 for NYUDv2. This value determines the maximum distance allowed between the predicted edge results and the ground truth in matching.

\subsection{Results}~\label{sec:results}
\begin{figure}[t]
    \centering
    \includegraphics[width=0.925\columnwidth]{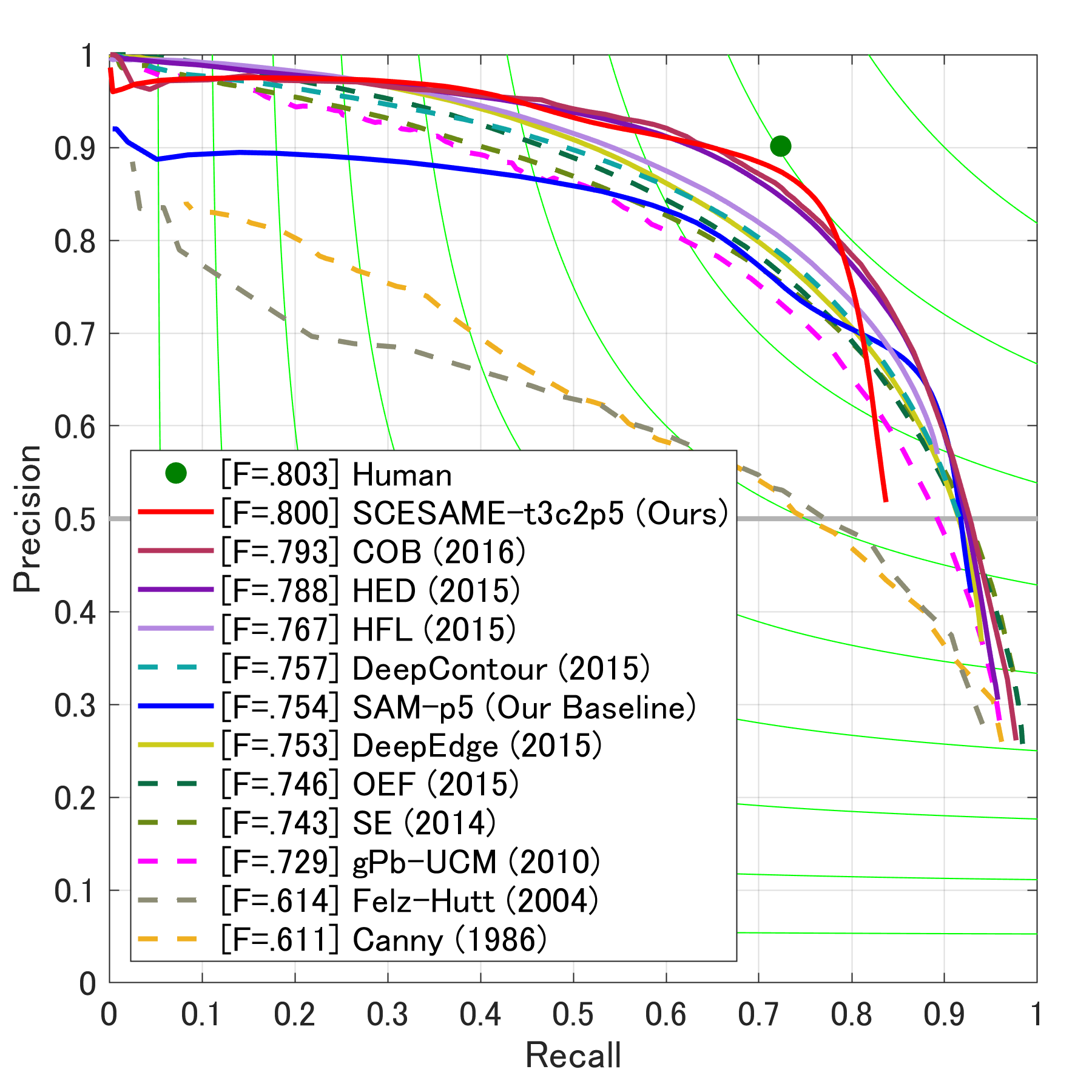}
    \caption{The precision-recall curves on BSDS500.}
    \label{fig:pr_bsds}
\end{figure}

\vspace*{-10pt}
\noindent
{\bf On BSDS500.}
We compare zero-shot edge detection with SCESAME to the following models:
human performance~\cite{DBLP:journals/corr/Kokkinos15},
traditional methods such as
Canny~\cite{DBLP:journals/pami/Canny86a}, 
Felz-Hutt~\cite{DBLP:journals/ijcv/FelzenszwalbH04}, 
gPb-owt-ucm~\cite{DBLP:journals/pami/ArbelaezMFM11}, 
SCG~\cite{DBLP:conf/nips/RenB12}, 
Sketch Tokens~\cite{DBLP:conf/cvpr/LimZD13}, 
PMI~\cite{DBLP:conf/eccv/IsolaZKA14}, 
SE~\cite{DBLP:journals/pami/DollarZ15}, 
OEF~\cite{DBLP:journals/corr/HallmanF14}, 
MES~\cite{DBLP:conf/iccv/SironiLF15},
as well as CNN-based models from 7-8 years ago including
DeepEdge~\cite{DBLP:conf/cvpr/BertasiusST15}, 
CSCNN~\cite{hwang2015pixelwise}, 
MSC~\cite{DBLP:journals/pami/SironiTLF16}, 
DeepContour~\cite{DBLP:conf/cvpr/ShenWWBZ15}, 
HFL~\cite{DBLP:conf/iccv/BertasiusST15}, 
HED~\cite{DBLP:conf/iccv/XieT15}, 
Deep Boundary~\cite{DBLP:journals/corr/Kokkinos15}, 
CEDN~\cite{DBLP:conf/cvpr/YangPCL016}, 
RDS~\cite{DBLP:conf/cvpr/LiuL16}, 
COB~\cite{DBLP:conf/eccv/ManinisPAG16}, 
and state-of-the-art methods such as
EDTER~\cite{DBLP:conf/cvpr/PuHLGL22}, 
UAED~\cite{Zhou_2023_CVPR}.
The results of these experiments are taken from previous works\cite{DBLP:journals/corr/Kokkinos15,DBLP:conf/cvpr/PuHLGL22,Zhou_2023_CVPR}. We also compare the original results of SAM~\cite{DBLP:journals/corr/abs-2304-02643}, our reimplementation of AMG, and the results of AMG using BZP.
For the hyperparameters $t$ and $c$ of TMS and SC, we considered the values $(t,c)=(2,2),(2,3),(3,2),(3,3)$. 
The results are presented in Table~\ref{tab:bsds}. For BSDS500, the best ODS and OIS are obtained at $(t,c) = (3,2)$. Edge detection with SCESAME surpasses traditional methods for both ODS and OIS. It outperforms most CNN-based methods from 7-8 years ago for ODS except Deep Boundary, and for OIS except Deep Boundary and COB. It also comes close to human performance. Compared to AMG, we observe improvements for both OIS and ODS. However, there is still a considerable gap in the results compared to the state-of-the-art methods.

For methods where results are available at different thresholds, the precision-recall curves are shown in Figure~\ref{fig:pr_bsds}. Edge detection with SCESAME achieves a high F-score (ODS) and is close to human performance. In Table~\ref{tab:bsds}, edge detection with SCESAME does not show such a high AP compared to ODS and OIS, and its cause, as seen in Figure~\ref{fig:pr_bsds}, is the lack of high recall values. We will discuss this further in \S~\ref{sec:discussion}.

Although there are discrepancies between the original SAM results and our reimplementation, the trends observed using AMG with BZP are consistent with the original. Therefore, we use it as the baseline for our experiments.

\paragraph{On NYUDv2.}
\begin{table}[t!]
\setlength{\abovecaptionskip}{0pt}
\centering
\footnotesize
\renewcommand\arraystretch{0.9}
\renewcommand\tabcolsep{5.0pt}
\begin{tabular}{@{\hspace{0.2em}}c@{\hspace{0.2em}}|l|c|ccc}
    \midrule
    \multicolumn{2}{c|}{Method} &\makecell[c]{Pub.'Year}    & ODS   & OIS   & AP        \\
    \midrule
    \multirow{7}{*}{\rotatebox{90}{Traditional}}
    & gPb-ucm~\cite{DBLP:journals/pami/ArbelaezMFM11}     & PAMI'11     & 0.632 & 0.661 & 0.562     \\
    & Silberman~\etal~\cite{DBLP:conf/eccv/SilbermanHKF12}  & ECCV'12     & 0.658 & 0.661 & -  \\
    & gPb+NG~\cite{DBLP:conf/cvpr/GuptaAM13}   & CVPR'13     & 0.687 & 0.716 & 0.629     \\
    & SE~\cite{DBLP:journals/pami/DollarZ15}            & PAMI'14     & 0.695 & 0.708 & 0.679     \\
    & SE+NG+~\cite{DBLP:journals/corr/GuptaGAM14}         & ECCV'14     & 0.706 & 0.734 & \TBEST{0.738}     \\
    & OEF~\cite{DBLP:journals/corr/HallmanF14}           & CVPR'15     & 0.651 & 0.667 & -         \\
    & SemiContour~\cite{DBLP:conf/cvpr/ZhangXSY16}       & CVPR'16   & 0.680 & 0.700 & 0.690   \\
    \midrule
    \multirow{6}{*}{\rotatebox{90}{CNN-based}}
    & HED~\cite{DBLP:conf/iccv/XieT15}       & ICCV'15     & 0.720 & 0.734 & 0.734     \\
    & RCF~\cite{DBLP:journals/corr/LiuCHWB16}       & CVPR'17     & 0.729 & 0.742 & -         \\
    & AMH-Net~\cite{DBLP:conf/nips/XuOARWS17} & NeurIPS'17  & \SBEST{0.744} & \SBEST{0.758} & \SBEST{0.765}     \\
    & LPCB~\cite{DBLP:conf/eccv/DengSLWL18}    & ECCV'18     & 0.739 & \TBEST{0.754} & -         \\
    & BDCN~\cite{DBLP:conf/cvpr/HeZYSH19}      & CVPR'19     & \BEST{0.748} & \BEST{0.763} & \BEST{0.770}     \\
    & PiDiNet~\cite{DBLP:conf/iccv/0002LYH00P021}& ICCV'21     & 0.733 & 0.747 & -         \\
    \midrule
    & SAM-p5 (Our Baseline)        & -      & 0.699 & 0.719  & 0.707 \\
    & SCESAME-t3c2p5 (Ours)           & - & \TBEST{0.742} & \TBEST{0.754} & 0.707 \\
    \midrule 
    \midrule
    & EDTER~\cite{DBLP:conf/cvpr/PuHLGL22} (SOTA)               & CVPR'22           & {\bf 0.774}  & {\bf 0.789}  & {\bf 0.797}\\    
    \midrule
  \end{tabular}
\caption{Results on NYUDv2~\cite{DBLP:conf/eccv/SilbermanHKF12} testing set. The notation t3c2p5 represents $t=3,c=2,p=5$ and so on. The best three results, excluding the SOTA method, are highlighted in \BEST{red}, \SBEST{blue}, and \TBEST{purple}. The SOTA method is highlighted in {\bf bold}.\\}
\label{tab:nyud}
\vspace{-14pt}
\end{table}
We also evaluated the performance on RGB images using the NYUDv2 dataset. Our comparison involved SCESAME-t3c2p5 and SAM-p5 against various models, including traditional methods such as gPb-ucm~\cite{DBLP:journals/pami/ArbelaezMFM11}, Silberman~\etal~\cite{DBLP:conf/eccv/SilbermanHKF12}, gPb+NG~\cite{DBLP:conf/cvpr/GuptaAM13}, SE~\cite{DBLP:journals/pami/DollarZ15}, SE+NG+~\cite{DBLP:journals/corr/GuptaGAM14}, OEF~\cite{DBLP:journals/corr/HallmanF14}, SemiContour~\cite{DBLP:conf/cvpr/ZhangXSY16}, and CNN-based models such as HED~\cite{DBLP:conf/iccv/XieT15}, RCF~\cite{DBLP:journals/corr/LiuCHWB16}, AMH-Net~\cite{DBLP:conf/nips/XuOARWS17}, LPCB~\cite{DBLP:conf/eccv/DengSLWL18}, BDCN~\cite{DBLP:conf/cvpr/HeZYSH19}, PiDiNet~\cite{DBLP:conf/iccv/0002LYH00P021}, and the state-of-the-art method EDTER~\cite{DBLP:conf/cvpr/PuHLGL22}. The experimental results are taken from previous work\cite{DBLP:conf/cvpr/PuHLGL22}. These results are presented in Table~\ref{tab:nyud}. Edge detection with SCESAME outperforms traditional methods and performs almost as well as CNN-based methods for ODS and OIS. 
Similar to the BSDS500 results, we observe an improvement over AMG for OIS and ODS. However, there remains a noticeable performance gap when compared with the state-of-the-art method.

Note that there are fewer methods tested on NYUDv2 compared to BSDS500, and CNN-based methods can further improve their performance by using both RGB and HHA images during training.

\subsection{Ablation Study}~\label{sec:ablation}
\begin{table}[t]
    \centering
    \footnotesize
    \begin{tabular}{l|@{\hspace{0.75em}}c@{\hspace{0.75em}}c@{\hspace{0.75em}}c@{\hspace{0.75em}}|@{\hspace{0.75em}}c@{\hspace{0.75em}}c@{\hspace{0.75em}}c@{\hspace{0.75em}}}
    \midrule
         Method & TMS & SC & BZP & ODS & OIS & AP \\
        \midrule
{\small SAM (Recalc.)} & &  &  & 0.730 & 0.754 & 0.729 \\
{\small SAM-p5} & &  & \checkmark & 0.754 & 0.779 & 0.763 \\
\midrule
{\small TMS-t3} & \checkmark &  & & 0.757 & 0.769 & 0.718 \\
{\small TMS-t3p5} & \checkmark &  & \checkmark  & 0.797 & 0.812 & \textbf{0.792} \\
\midrule
{\small SC-c2} & & \checkmark &  & 0.743 & 0.762 & 0.731 \\
{\small SC-c2p5} & & \checkmark & \checkmark  & 0.771 & 0.792 & 0.773 \\
\midrule
{\small SCESAME-t3c2} & \checkmark & \checkmark & & 0.753 & 0.764 & 0.693 \\
{\small SCESAME-t3c2p5} & \checkmark & \checkmark & \checkmark & \textbf{0.800} & \textbf{0.814} & 0.773 \\
    \midrule
    \end{tabular}
    \caption{Ablation study on BSDS500~\cite{DBLP:journals/pami/ArbelaezMFM11} testing set. The notation t3c2p5 represents $t=3,c=2,p=5$ and so on.}
    \label{tab:ablation}
\end{table}
As seen in \S~\ref{sec:proposed}, TMS, SC, and BZP are independent processes. Therefore, an ablation study is performed on BSDS500 using the parameters that gave the best ODS and OIS performance: $t=3$, $c=2$, and $p=5$. The experimental results are presented in Table \ref{tab:ablation}. It is evident that the performance improves when BZP is used in AMG, TMS, SC, and SCESAME. While TMS-t3p5 outperforms SC-c2p5, SCESAME-t3c2p5 shows superior ODS and OIS compared to TMS-t3p5. This suggests the importance of combining TMS and SC. 
Note that for $t=3$ and $c=2$, the proportion of mask removal and mask combination is different and that TMS-t3p5 had the highest AP among them. 

\section{Discussion}\label{sec:discussion}
In this section, we first discuss the limitations of edge detection with SCESAME based on the results in \S~\ref{sec:results}. Then we give some suggestions for future work and conclude.

\paragraph{Limitation.}
Edge detection with SCESAME shows a gap compared to state-of-the-art methods. The lower AP compared to ODS and OIS can be attributed to the suppression of edges during mask removal and combination. While BZP is effective, it can also fill true positive edges with zeros, contributing to low recall.
AMG may be more practical than SCESAME for detecting finer edges. In addition, the datasets we used are based on a few annotations, so the edges removed by SCESAME (or overdetected by AMG) are not necessarily redundant.

\paragraph{Future Work.}
Instead of TMS, selection based on mask importance, random selection, or selection from mask features can preserve even small masks that are critical for edge detection. 
Affinity in (\ref{eq:scesame}), based on mask position and overlap ratio, is a simple model with room for improvement. 
The parameters $t,c,p$ in TMS, SC, and BZP are currently fixed, so choosing optimal values for each image may be beneficial. 
Consideration of a few-shot fine-tuning model could further improve its performance. 
In this study, SCESAME was used for edge detection, but if AMG is used in downstream tasks, SCESAME can also be used.

\paragraph{Conclusion.}
This paper proposes a novel zero-shot edge detection with SCESAME based on AMG. This method, which consists of three steps, overcomes the overdetection problem of AMG edges. Experimental results on the BSDS500 and NYUDv2 show that despite being a simple zero-shot method, our approach exhibits performance comparable to human performance and recent CNN-based methods. 
These results indicate that our method effectively enhances the utility of SAM and can be a new direction in zero-shot edge detection methods.

\clearpage

{\small
\bibliographystyle{ieee_fullname}
\bibliography{custom}
}

\end{document}